\documentclass[runningheads]{llncs}
\usepackage{graphicx}
\usepackage{amsmath,amssymb} 
\usepackage{color}
\usepackage[width=122mm,left=12mm,paperwidth=146mm,height=193mm,top=12mm,paperheight=217mm]{geometry}
\usepackage{times}
\usepackage{epsfig}
\usepackage{graphicx}
\usepackage{amsmath}
\usepackage{amssymb}
\usepackage{graphicx}
\usepackage{setspace}

\usepackage{multirow}
\usepackage{booktabs}

\usepackage{calrsfs}
\DeclareMathAlphabet{\pazocal}{OMS}{zplm}{m}{n}

\begin{document}
\pagestyle{headings}
\mainmatter

\title{Weakly Supervised Object Discovery by \\ Generative Adversarial \& Ranking Networks} 

\titlerunning{A very long title}

\authorrunning{authors running}

\author{Ali Diba$^{1}$, Vivek Sharma$^{2}$, Rainer Stiefelhagen$^{2}$, and Luc Van Gool$^{1,3}$}


\institute{$^{1}$ESAT-PSI, KU Leuven, $^{2}$CV:HCI, KIT, Karlsruhe, and $^{3}$CVL, ETH Z\"{u}rich \\ 
$^{1}$\{firstname.lastname\}@esat.kuleuven.be, $^{2}$\{firstname.lastname\}@kit.edu}

\maketitle

\begin{abstract}
The deep generative adversarial networks (GAN) recently have been shown to be promising for different computer vision applications, like image editing, synthesizing high resolution images, generating videos, etc. These networks and the corresponding learning scheme can handle various visual space mappings. We approach GANs with a novel training method and learning objective, to discover multiple object instances for three cases: 1) synthesizing a picture of a specific object within a cluttered scene; 2)  localizing different categories in images for weakly supervised object detection; and 3) improving object discovery in object detection pipelines. A crucial advantage of our method is that it learns a new deep similarity metric, to distinguish multiple objects in one image. We demonstrate that the network can act as an encoder-decoder generating parts of an image which contain an object, or as a modified deep CNN to represent images for object detection in supervised and weakly supervised scheme. Our ranking GAN offers a novel way to search through images for object specific  patterns. We have conducted experiments for different scenarios and demonstrate the method performance for object synthesizing and weakly supervised object detection and classification using the MS-COCO and PASCAL VOC datasets.
\end{abstract}
\section{Introduction} \label{sec:intro}
Discovering objects in scenes is one of the fundamental problems in the computer vision field. Deep neural networks have been promising for the purpose, but still need a large-scale, annotated dataset for training. There have been numerous efforts to work with an unsupervised setup, e.g. based on generative or, most notably, mixed generative-discriminative networks (GANs)~\cite{goodfellow,perarnau,videogan}. Some recent works have focused on extending the training dataset, using a supervised dataset to synthesize additional, realistic data~\cite{dosovitskiy2,nguyen}. Powered by adversarial learning, supervised GANs can generate more accurate images based on distinguishing different classes and fake from real data.

\begin{figure}[t]
\centering
{\includegraphics[width=0.75\columnwidth]{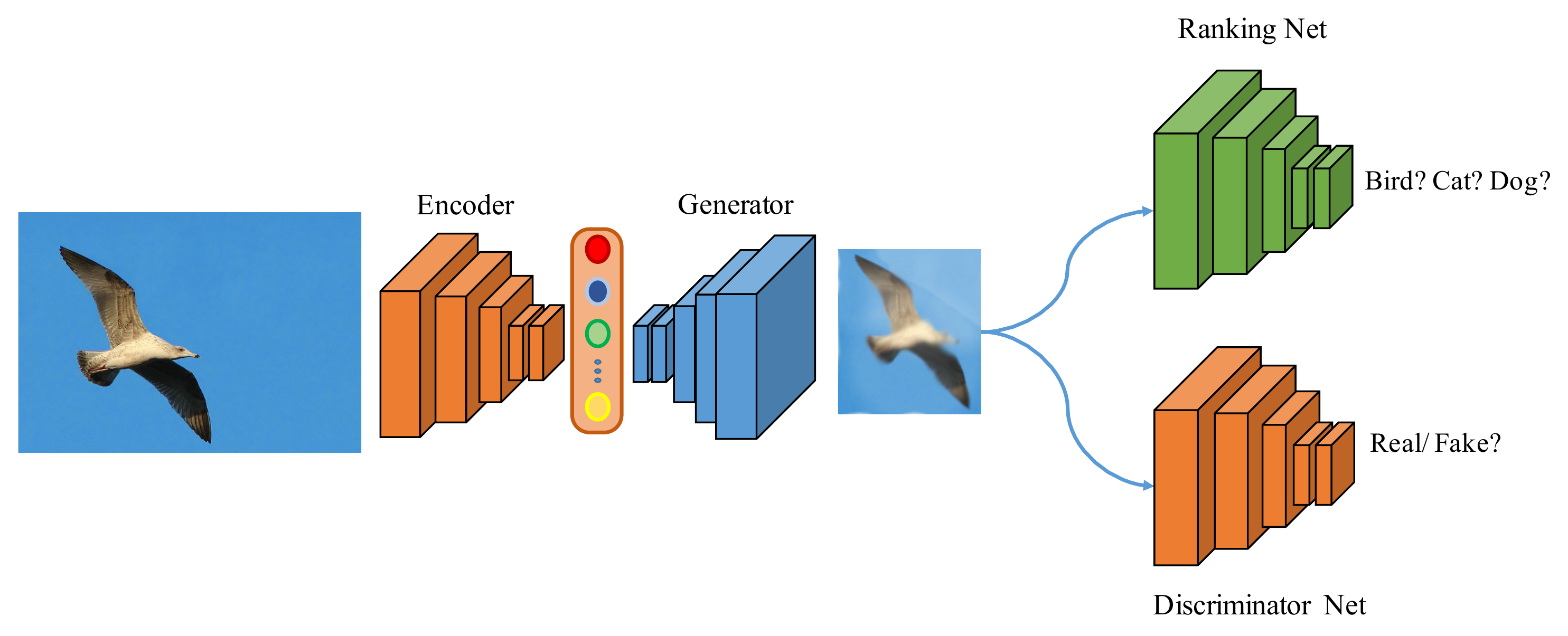} 
} 
\caption{The proposed generative ranking and adversarial networks take an image containing different object categories and synthesize a specific object instance among the other objects in a scene. The proposed method allows generative models to tackle recognition and detection problems, and is validated for weakly supervised object detection.
} 
\label{fig:front}
\end{figure}

To tackle the weakly supervised object detection, creating accurate object templates for each image independently and finding location of object instances by those templates is one solution. To the best of our knowledge, our work is the first to tackle object detection and discovery using GANs, and that also in a weakly supervised manner. As a wider context for the work, using a combination of encoder based conditional GANs with ranking objective can be utilized to extract the correct templates of objects in an image. The image is encoded by a CNN and our generative network draws a realistic sample of a specific object in the image using indication of its location. The discriminator and our ranking network help the generator to synthesize the most realistic and correct sample. Our scheme combines conditional GANs, adversarial learning, and a feature space similarity metric. The objective, i.e. the new ranking loss, helps the generator at training time to discriminate between the objects, and to learn the relevant features from the encoder network that allows it to draw the object samples (see Fig.~\ref{fig:front}). Thus our motivation to propose such a pipeline is to solve the problem of finding nice object instance templates for each image and to do so powerful and novel generative models are promising solutions. 

Nguyen et al.~\cite{nguyen} showed that adding supervision information like class labels helps to learn generating more realistic images. In addition to using object labels, we incorporate a similarity ranking between different category of samples. This strengthens the cooperation between the generator and discriminator when synthesizing the object instances.

Another advantage of our work is that it exploits deep generative models to allow for weak supervision. For training the object detector in weakly labeled setting, we do not use ground-truth bounding boxes. Instead, the weakly supervised object detector searches itself for the possible locations of object samples in the training images. Our novel GAN generates object templates similar to the actual object samples present in the image. After this step, it uses the synthesized templates which look like the real objects to find the accurate location of object instance. Eventually these locations are handed over to train the detector as a pseudo ground-truth. Our GAN and ranking networks use ImageNet-like categories without bounding boxes to learn feature space similarity and supervise the ranking process.

Our approach has been evaluated in several scenarios: realistic image generation, object instance extraction, multi-class object classification, and object localization with single/multi instances per image. Our contributions are as follows:
\begin{itemize}
  \item Presenting a new approach for supervised learning of generative ranking \& adversarial networks to synthesize realistic objects.
  \item Proposing a GAN with multiple objective losses to create accurate samples of objects in an image and to draw realistic-looking images.
  \item Training a weakly supervised generative network as a solution to weakly supervised object detection and training with hallucinated samples.
\end{itemize}
The rest of the paper is organized as follows. Section~\ref{sec:related} discusses related work. We introduce the GANs in section ~\ref{sec:gan} and  explain the core of the work in section~\ref{sec:method} and describe the weakly supervised GAN for object detection in section~\ref{sec:weakly}. The experiments are presented in section~\ref{sec:exp}. Finally,  in section~\ref{sec:conc} conclusions are drawn. 


\section{Related Work} \label{sec:related}

\textbf{Generative Models:} Recently several attempts have been made to improve image generation using generative models. The most popular generative model approaches are Generative Adversarial Networks (GANs)~\cite{goodfellow},  Variational Autoencoders (VAEs)~\cite{kingma2013auto}, and Autoregressive models~\cite{oord2016pixel}. And their variants, e.g. conditional GANs, Invertible Conditional GANs, Deep Convolutional GANs (DC-GANs)~\cite{radford},  etc. Radford et al.~\cite{radford} use a Conv-Deconv GAN architecture  to learn good  image representation for several image synthesis tasks.  Denton et al.~\cite{denton} use a Laplacian pyramid of generators and discriminators to synthesize multi-scale high resolution images. Mirza and Osindero~\cite{cgan} train GANs by explicitly providing a conditional variable to both the generator and the discriminator, using one-hot encoding to control generated image features, namely conditional GANs (cGANs). Reed et al.~\cite{reed} use a DC-GAN conditioned on text features encoded by a hybrid character-level convolutional RNN. Perarnau et al.~\cite{perarnau} use an encoder with a cGAN, to inverse the mapping of a cGAN for complex image editing, calling the result Invertible cGANs. Dumoulin et al.~\cite{dumoulin} and Donahue et al.~\cite{donahue} use an encoder with GANs. Makhzani et al~\cite{makhzani} and Larsen et al.~\cite{larsen} use a similar idea to~\cite{perarnau}, but combining a VAE and GAN to improve the realism of the generated images.

Also with GANs an appropriate reconstruction loss is necessary to avoid blurry results, because the distances are computed in the image space. Solutions have been proposed to mitigate this problem. One is to measure the similarity in the feature space, instead of the image space, as proposed in~\cite{larsen}. Similarly in~\cite{dosovitskiy2}, Dosovitskiy et al. use perceptual similarity metrics between image features. This results in  sharp and realistic generated images.

Several applications have been based on different types of GANs. Mathieu et al.~\cite{mathieu} predict future frames in videos, conditioned on previous frames. Larsen et al.~\cite{larsen} generate realistic images of faces. Dosovitskiy et al.~\cite{dosovitskiy} and Rifai et al.~\cite{rifai} generate images of object categories given high-level information about the desired object. Reed et al.~\cite{reed2} generate realistic images from text and landmarks. Pathak et al.~\cite{pathak} use context encoders to generate the contents of an arbitrary missing image region conditioned on its surroundings. Isola et al.~\cite{isola} learn the mapping from input images to target images. Nguyen et al.~\cite{nguyen} generate high-resolution, photo-realistic images using text-to-image generative models. Wang et al.~\cite{cite39} generate images from the surface normal map. Zhu et al.~\cite{cite49} modify the appearance of an image while preserving realism, guided by user constraints. Zhou et al.~\cite{cite48}  create depictions of objects at future times in time-lapse videos. Li et al.~\cite{cite25} efficiently synthesize textures for style transfer. Yoo et al.~\cite{cite43} show pixel-level domain transfer to generate realistic target images.

\textbf{Weakly supervised object detection}

$-$\textbf{Weakly supervised learning:} Over the last decade, several weakly supervised object detection methods have been studied that were using multiple instance learning (MIL)~\cite{bilen14,bilen15,siva,song14}. Because of its non-convexity, MIL tends to get stuck in local optima, thus making it dependent on the initialization of object proposal instances in the positive/negative bags. To alleviate this shortcoming, many proposed strategies have been seeking a better initialization~\cite{does10,Siva12,siva,song14a} or have focused on regularizing the optimization strategies~\cite{bilen14,bilen15,cinbis}. The majority~\cite{reed14,sukh14} used large and noisy collections of object proposals to train the object detectors. We took a new way and approach it via generative networks to deal with weakly supervised object detection. In our work, we show that providing a level of supervision to the generative networks is beneficial for an accurate object localization. 

$-$\textbf{CNN based weakly supervised object detection:} Recently, several efforts have been made to let CNNs classify objects with weak supervision \cite{Contextlocnet,bilen16,li16}. Oquab et al.~\cite{Oquab14} compute a mid-level image representation for feature discrimination, employing a pre-trained CNN. In~\cite{laptev15}, the same authors modify the CNN architecture to coarsely localize object instances in images, thus improving the classification performance.  Bilen et al.~\cite{bilen16} use CNNs to operate at the level of image regions, simultaneously selecting regions and classifying. Li et al.~\cite{li16} address the problem via progressive domain adaptation for joint  classification and detection, using a pre-trained CNN network.  Diba et al.~\cite{diba} proposed cascaded stages for both object proposal and detection using multiple loss functions.

The main distinction between the aforementioned work and ours is that we use GANs for weakly supervised object detection. To the best of our knowledge it is the first end-to-end network initializing the location of objects using GANs for such task. Furthermore, we introduce generative ranking and adversarial networks for the realistic synthesis of objects.


\section{Generative Adversarial Networks} \label{sec:gan}

Generative adversarial networks normally include two parts: a generator network $G$ and a discriminator network $D$. The generator tries to beat the discriminator in a min-max game to distinguish between real and fake images. If the network was perfectly modeled, the generator will have learned the distribution of $p_{data}$ (real distribution of real images) so that it can deceive the discriminator network in believing synthesized patterns to be real ones. The formulation of the min-max game is:
\begin{equation}
\begin{split}
\min_{G}\,\max_{D}\, v(G,D) = E_{x\sim p_{data}(x)}[log D(x)] + \qquad \\
E_{z\sim p_{z}(z)}[log(1-D(G(z)))]
\end{split}
\label{eq:1}
\end{equation}
where $z$ is an input vector drawn from the distribution $p_{z}$. It can be a noisy sample~\cite{dosovitskiy,goodfellow} or a feature vector representation~\cite{nguyen}. When the game reaches convergence $p_{z}$ equals $p_{data}$.  

\subsection*{Conditional GAN}
Conditional GANs~\cite{cgan,perarnau} were introduced as a more sophisticated extension of GANs, to widen their applicability. These networks are useful for cases like image editing.  Formally, we have:
\begin{equation}
\begin{split}
\min_{G}\,\max_{D}\, v(G,D) = E_{x,y\sim p_{data}(x)}[log D(x,y)] + \qquad \\
E_{z\sim p_{z}(z),y'\sim p_{y}}[log(1- D(G(z,y'),y'))]
\end{split}
\label{eq:1}
\end{equation}
One can apply different conditions $y$ to control the generator when training a cGAN.  Our method uses a feature representation extracted from images by an encoder network (CNN) and combined with a location feature, similar to \cite{dosovitskiy2,pathak}, as the condition for training. This feature vector needs to be embedded into the input of our GAN in order to obtain the distribution of the data.  

\begin{figure*}[t]
 \centering
 \includegraphics[width=1\columnwidth]{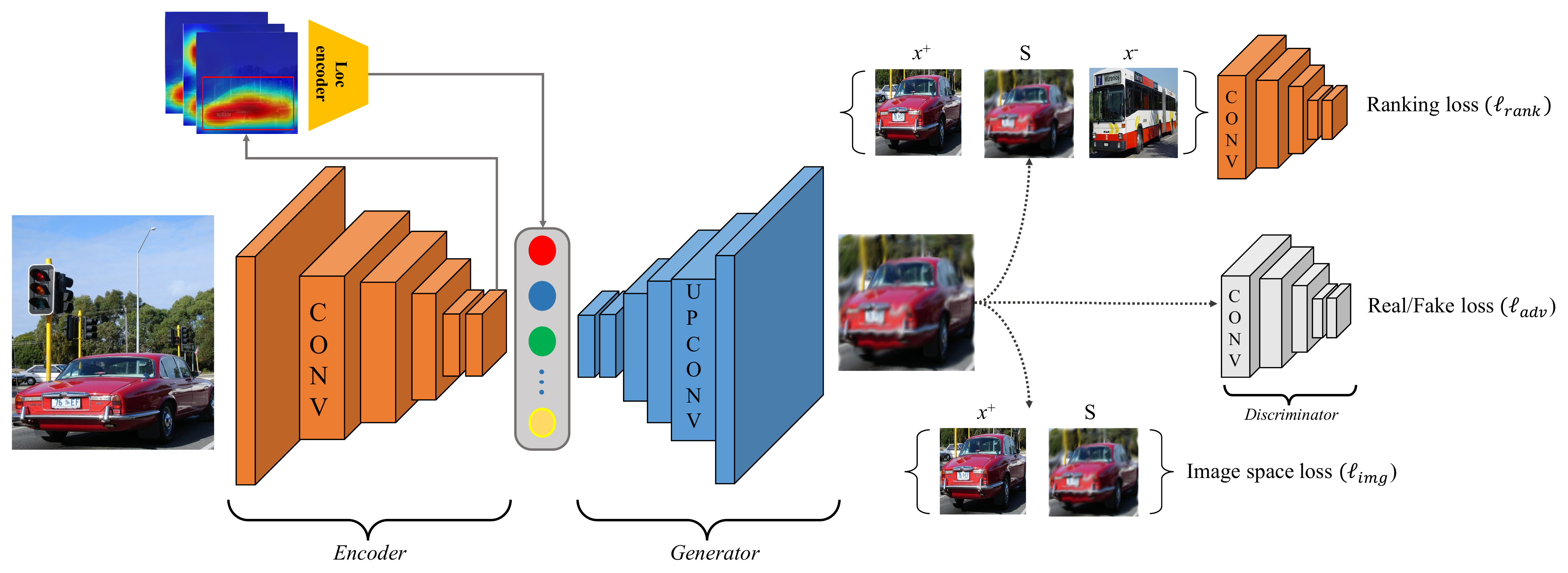}
 \caption{The end-to-end training pipeline of our generative network, using different types of losses. \textbf{1)} The encoder network produces the input of the generator. \textbf{2)} The generator network synthesizes an object instance. \textbf{3)} The synthesized object image is fed to the discriminator block,  which is composed of ranking, adversarial and image space objectives. This joint objectives help to improve the generation and discrimination capabilities of the whole network.}
\label{fig:pipeline}
\end{figure*}

\section{Object Discovery by GAN} \label{sec:method}

In this section, we present our proposed pipeline, i.e. a new deep generative network which is able to localize different object samples and to draw them at high resolution on demand. It creates templates identical to the real objects in an image, where multiple categories may be present simultaneously. The main motivation of this work is to effectively localize objects in a scene with or without using any ground-truth object locations in the training phase of the deep generative networks. We show that our new training pipeline of generative networks is capable of object discovery in complex scenes. Since the main application of this work is its deployment for  weakly supervised object detection, we believe our proposed method can help the task to find accurate object locations for final detector training. We show through our extensive experiments that the new proposed generative model can localize the object samples effectively, and thus the synthesized instances improve the detection pipeline by a significant margin. Here, we firstly discuss the feasibility of our method to synthesize object instances in a supervised setting and later we employ this solution on weakly supervised task.

Figure~\ref{fig:pipeline} illustrates the pipeline of our proposed architecture. Our architecture has three components: $\bf{1)}$ a visual and location encoder, $\bf{2)}$ a generator network (can be considered as a decoder), and $\bf{3)}$ a discriminator and ranking networks. The encoder extracts a high level visual representation of an input image. Afterwards a regional condition (discussed later) is added to the encoded features as input for the generator. The generator takes the entire encoded vector and produces the object it believes to be present in the specified region. This part provides information about the visual distribution of the image and about the location that we are interested in. After a synthetic image (object instance) is produced, the discriminator (with our novel configuration and set of losses) keeps correcting the generator in two ways: (i) by evaluating the similarity between the created sample and the actual object instance vs another object category, through the ranking network; and (ii) by maximizing the benefits from adversarial objective, through the discriminator network. We also show that image space loss which minimizes the difference between the synthesized object image and the real object instance is beneficial. We use the pre-trained GAN model from \cite{dosovitskiy2}, which provides a generative model to create realistic, synthetic images by perceptual similarity metric learning. So in this way, we handle object instance discovery by a generative network.

\subsection{Encoder network}
The encoder network in our pipeline is convolutional network, like AlexNet~\cite{alexnet} or VGG-16~\cite{vgg} trained on the ImageNet dataset~\cite{imagenet} and in fully supervised cases fine-tuned on a target dataset of objects (e.g. PASCAL or COCO) which is not the same case in the weakly supervised method. We use AlexNet or VGG, both trained by global average pooling~\cite{gap}, but replace their fully connected layers by a new one, inspired by~\cite{pathak}. The last fully connected layer carries the object activation neurons, after the $conv$ layers. In the case of AlexNet, the size of this layer is 6400 $(5\times5\times256)$. The weights of this network are updated during training, to improve the encoder's ability to distinguish separate object instances.

\textbf{Location encoding (Loc encoder):}
As shown in Fig.~\ref{fig:pipeline}, the class activation map (or heat map) of objects is fed to a small network with a convolution and a pooling layer, followed by a fully connected layer to encode the location. This encoded feature is used as the location condition for generative network. The location encoded features size is 320.
 
As the final input to the generative network, we combine the features that were obtained from the visual encoder with the location feature $(6400+320)$.  

\subsection{Generative network}
The generative part of network synthesizes object instances based on the encoded features, obtained from the previous stage instead of noise input as in regular GANs. Inspired by ~\cite{dosovitskiy2,gap}, our generator network contains multiple up-convolutional (deconvolutional) layers, which were pre-trained based on~\cite{dosovitskiy2}. This decoder focuses on a specified image region and draws the existing object instance.

\subsection{Discriminator-Ranking networks}
Most of the novelty in our work comes with this block. It evaluates the object instance synthesized by the generator to improve both the system's generating and discriminating ability. We propose to use multiple new losses to do the task in the discriminator section. The combined objective functions (discussed further) together learn to capture the most important visual relations within and between object classes in the image. It empowers the generator to draw more accurate object instances. Exploiting the supervision that comes with bounding boxes, the ranking loss can visually compare the synthesized sample with its instance of origin and another object instance in the image. This loss function helps to discriminate between the samples in the encoding feature space. This term increases the probability of creating samples that activate different neurons for different categories. The discriminator part of the pipeline has two separate branch: one to calculate the ranking loss and one for adversarial training.

\paragraph{Ranking Network}
Given the synthesized object instance $S$, the original object instance $x^+$ and another object instance $x^-$ from a different category, we have $(S,x^+)$  as a `positive pair' and $(S,x^-)$ as a `negative pair'. As an example: suppose an image shows a cat and a dog, and we want to draw the cat instance only, so the cat is the positive and the dog is the negative sample. Using these two pairs, the ranking loss trains with a new similarity metric to improve the generative model. This network is identical to the encoder and has the same weights. We extract the features $f$ from the last layer of ranking network (the network with ranking loss in Fig.~\ref{fig:pipeline}) to calculate the loss using the cosine distance. We have:

\begin{equation}
\begin{split}
Dist(S,x^+) = 1 - \frac{f(S).f(x^+)}{\|f(S)\| \|f(x^+)\|}
\end{split}
\label{eq:1}
\end{equation}
The concept learned by ranking amounts to reducing the distance between the synthesized image and the positive sample, and making it smaller than the distance between the synthesized image and the negative sample. The objective function makes the feature space of similar objects correspondent. The hinge loss for the ranking task is given as:
\begin{equation}
\begin{split}
\ell_{rank} = \max\{0, Dist(S,x^+) - Dist(S,x^-)\}
\end{split}
\label{eq:1}
\end{equation}
\noindent Some related works~\cite{dosovitskiy2,nguyen} use feature space loss that helps to minimize the feature distance between synthesized and real images. Using the proposed ranking loss overcomes the need for such a feature space loss in the condition when we have bounding box of object during training.

\paragraph{Adversarial Network}
In our pipeline, the adversarial training helps to prevent the network from producing blurred images and converge to much more realistic object instances. The discriminator network $D$ tries to distinguish between fake versus real images, and is trained by minimizing the objective function given by:
\begin{equation}
\begin{split}
\ell_{disc} = -\sum_{i}{log(D(x_{i}^+)) + log(1-D(S_{i}))}
\end{split}
\label{eq:1}
\end{equation}
and the generator minimizes the objective function, given as:
\begin{equation}
\begin{split}
\ell_{adv} = -\sum_{i}{log(D(S_{i}))}
\end{split}
\label{eq:1}
\end{equation}
The discriminator network is composed of 5 convolutional and two fully connected layers with softmax layer for fake/real prediction. The network is pre-trained based on regular GAN training from~\cite{dosovitskiy2}.

\paragraph{Image loss.}
This loss optimizes the generator's input vector (i.e. the output of Encoder and Loc encoder) to produce object samples that can stimulate the class activation to distinguish between objects effectively. Additionally it also provides solid gradients to fix unstable behavior of adversarial training. The following function should be maximized, given as:
\begin{equation}
\begin{split}
\ell_{img} = \sum_{i}{\|S_{i} - x_{i}^+\|^2_2 }
\end{split}
\label{eq:1}
\end{equation}

Using such a image objective function along with adversarial and ranking helps to improve the image reconstruction ability of the pipeline.

\paragraph{End-to-End training.}
The whole encoding-generating pipeline, with its three objective functions, is learned jointly by end-to-end stochastic gradient descent optimization. The total loss function of the network is given as:
\begin{equation}
\begin{split}
\ell_{Total} =  \alpha_{rank}\ell_{rank} +  \alpha_{img}\ell_{img} + \alpha_{adv}\ell_{adv}  
\end{split}
\label{eq:totalloss}
\end{equation}
\noindent We provide more details about the training parameters such as loss coefficients $\alpha$'s in the experiments section.


\section{Weakly Supervised GAN} \label{sec:weakly}

Here, we propose weakly supervised object detection using our proposed encoder-decoder set of networks, without bounding boxes for training. Hence, instead we need a method to discover the most probable locations of objects for training. Since, we do not use object ground-truth bounding boxes (weakly supervised condition) in our ranking network, that means the object samples are not available to calculate the ranking objective function, for this reason we use the similar categories of objects from ImageNet dataset.

In preparation $x^+$ for a category like cat, we pick up a random image from the same category of ImageNet, and $x^-$ is taken from different category of ImageNet like dog. So we have the positive and negative samples to evaluate the synthesized cat using the ranking loss.

Since we do not have a good reference to calculate the image objective function, the feature space loss is added to the network to boost object instance localization for generating samples. This loss is a complement to the ranking loss in the case of weak supervision, as real samples can not be used to rank the synthesized images. Consequently, we have to modify the way in which we use the feature space loss in the current situation. The object proposals \cite{SS,edgebox} are extracted and based on the resulting object heat map (similar to the supervised setup), we choose the top K=5 boxes with the highest heat score. To address the feature space loss, we average the features extracted from these five boxes as global representation of the possible object instance. And the averaged feature vector is applied to the feature space loss. The loss for features is computed as:
\begin{equation}
\begin{split}
f_{avg}^+ = \sum_{j}{f(x_{ij}^+)} / K \qquad
\\
\ell_{feat} = \sum_{i}{\|f(S_{i}) - f_{avg}^+\|^2_2 }
\end{split}
\label{eq:1}
\end{equation}

\noindent The whole network is trained with the total loss of Eq.~\ref{eq:totalloss}, but we replace the image loss by the feature loss.

\begin{figure}[t]
\centering
{\includegraphics[width=0.8\columnwidth]{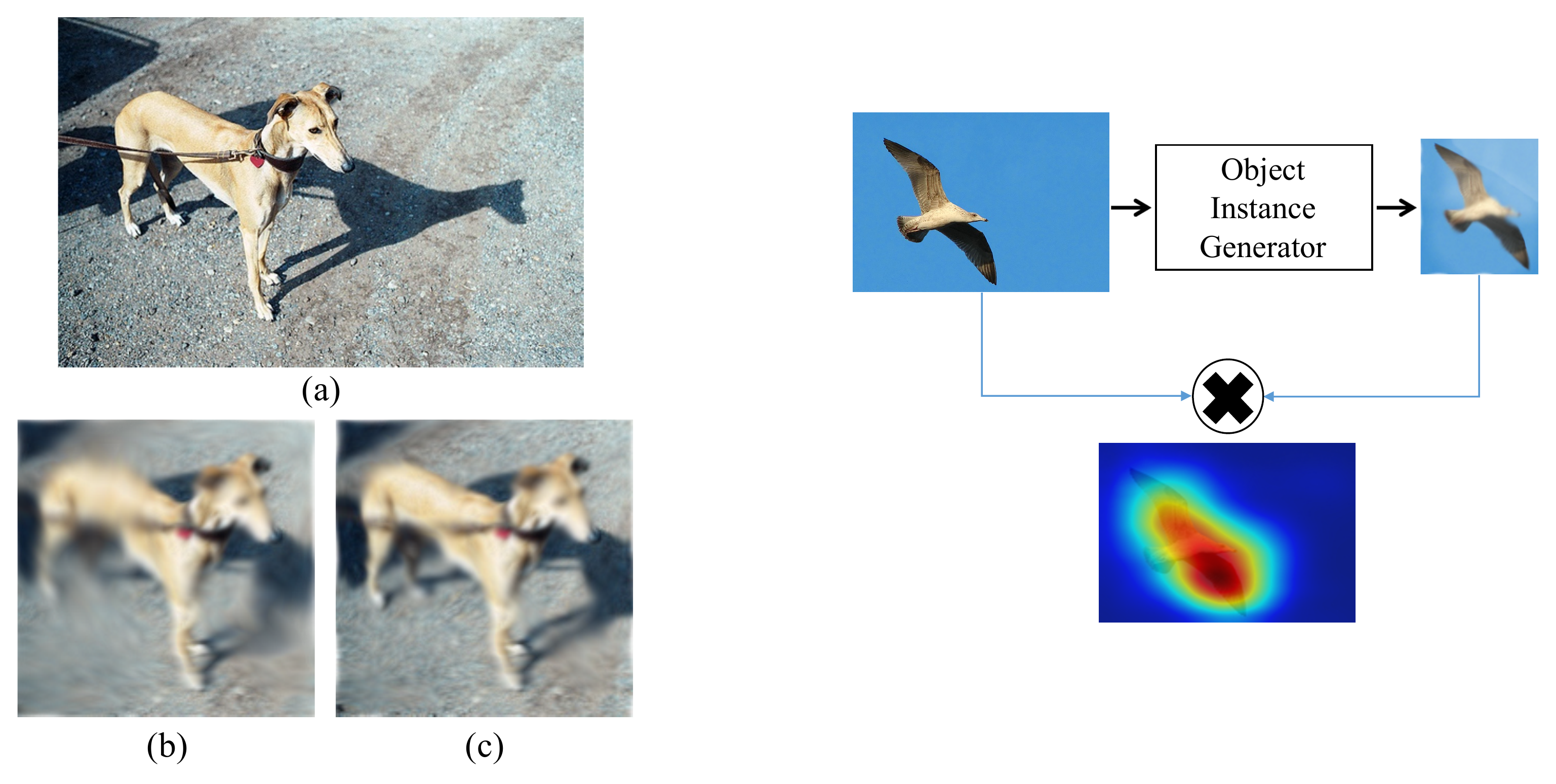} 
} 
\caption{\textbf{Left:} The figure shows an example of synthesizing the object in image (a) without/with using boxes in (b) and (c). \textbf{Right:} Finding an object location via template matching across the entire image, using the synthesized image as an effective template.} 
\label{fig:tmpmatch&sample}
\end{figure}

After the training, we can synthesize realistic object instances as they exist at different image locations. These synthesized object instances can be used as the likely object templates to accurately localize real objects in the input image. As an example in Figure~\ref{fig:tmpmatch&sample}, we compare the difference between synthesized images in supervised and weakly supervised methods. 
\noindent To extract the bounding box location, we run a simple template matching convolving at 3 different scales to create an accurate heat map as illustrated in Figure~\ref{fig:tmpmatch&sample}. Once all of the object samples are extracted, they can be used as pseudo ground-truth. So we train a state-of-the-art object detector like fast RCNN or SSD \cite{fastRCNN,SSD} with our pseudo ground-truth object bounding boxes. The experiments section shows our results, which outperform other weakly supervised methods on PASCAL VOC and MS-COCO. In some of the images, there exists multiple samples of the same object category. Similar to~\cite{gap}  where they apply connected-component method on the object heat-map to separate the samples, we also employ the same technique as a pre-processing step. Thus, for the model training we treat each of these samples as an independent instance.

\paragraph*{Synthetic objects for detection.} Another idea which improves the weakly supervised object detection, is to use synthetic data~\cite{3dSynth} or hallucinated samples. Based on our model we obtain a set of synthetic images which look quite similar to object instances and thus is used as an extra image resource. This is shown to be effective in our weakly supervised object detection experiments. In future work, we plan to investigate it further. More details are provided in the experiment section.


\section{Experiments}\label{sec:exp}

In this section, we discuss the experimental details and evaluations of our proposed method. First, we introduce the datasets on which the evaluations were performed. In our experiments, we have two main evaluations: (i) evaluating the capability of our model to generate single object instance from multiple objects in an image; and (ii) evaluating the effectiveness of our weakly supervised object detection.

\subsection{Implementation details}
Our implementation is done in Torch and uses some pre-trained models from Caffe implementation of ~\cite{dosovitskiy2}. All the networks are trained on two TitanX GPUs. For the network optimization, we use ADAM solver.

\subsection*{Architectures:}
\paragraph*{Encoder:} In our experiments, we tested AlexNet and VGG-16 networks as image encoder to prepare the input to the generator. These networks are pre-trained on the ILSVRC'12 ImageNet dataset. In the fully supervised task, they were fine-tuned on the target datasets PASCAL and COCO for classification. These networks were trained with global average pooling to improve their localization ability, as shown in~\cite{diba,gap}.

\paragraph*{Generator:} The configuration for this network is based on~\cite{dosovitskiy2} which has 3 fully connected layers initially, 6 up-convolution layers and also 3 convolution layers between the $up\-convs$.

\paragraph*{Discriminator block:} This part has two network components. The network for the ranking loss is identical to the encoder and has shared weights. The discriminator network with adversarial loss has 5 layers of convolution and thereafter a global average pooling, followed by two fully connected layers. A drop-out of 0.5 is applied before the $fc$ layers.

\paragraph*{Training details:} To train the total loss, we set the coefficients of the losses as: $\alpha_{rank} = 0.05$, $\alpha_{img} = 10^{-6}$ , $\alpha_{feat} = 10^{-5}$ and $\alpha_{adv} = 100$. For the Adam optimization parameters, we set the $\beta{1} = 0.9$, $\beta{2} = 0.99$ and start with a learning rate of 0.0001.

\subsection{Datasets and evaluation metrics}

The two datasets that we targeted for our evaluations are the PASCAL VOC 2007, 2010 and MS-COCO object datasets. The VOC datasets contain 20 categories, with almost 5K images for training and validation. The MS-COCO dataset is larger, including 80 categories with 80K training images, 40K validation images, and 20K testing images.
\paragraph*{Experimental metrics:} For the object detection evaluation, average precision (AP) and correct localization (CorLoc) have been used. Average precision is the standard metric as used in the PASCAL VOC competition. It takes a bounding box as a true detection when it has an intersection-over-union (IoU) of more than 50$\%$ with the ground-truth box. The \textit{Corloc} is the fraction of images for which the method obtained the correct location for at least one object instance, cf. MS-COCO. Average precision is also used for the classification evaluation.

For quantitative evaluation of synthesized images, we measure pixel-level similarity error i.e. RMSE and SSIM. The similarity error is calculated between the generated object instance and the real object.

\begin{table}[htb]
{\small
\tabcolsep=0.3cm
\caption{Quantitative performance: object synthesis on 1000 samples of PASCAL VOC objects.}
\label{tab:1}
\begin{center}
\begin{tabular}{lcc}
\toprule 
Method & SSIM & RMSE \\
\midrule
Dosovitskiy et al.~\cite{dosovitskiy2} & 0.16& 0.36 \\
Ours\_AlexNet & 0.25 & 0.30   \\
\textbf{Ours\_VGG} & \textbf{0.28} & \textbf{0.27}   \\
\midrule
Ours (whole\_img, box) & 0.22 & 0.32   \\
Ours (whole\_img) & 0.20 & 0.34   \\
\bottomrule
\end{tabular}
\end{center}}
\vspace{-1cm}
 
\end{table}

\subsection{Synthesizing quantitative evaluation}

Since there has been no other work very similar to ours, we have designed a new experiment on the PASCAL VOC dataset. We have evaluated the synthesizing power to produce object instances from test sets with 1000 samples. The object instances were cropped from the original images and we then generate the synthesized version of them using our pipeline and~\cite{dosovitskiy2}. We also measure the synthesis performance when other objects are present in the scene. In such a case, we have results with and without bounding box positions. Table~\ref{tab:1} summarizes the quantitative evaluations. It is shown that under the same conditions as used by~\cite{dosovitskiy2}, our network can reproduce images with a higher quality.

\subsection{Ablation study of losses}
We have evaluated different combinations of losses, with results shown in Table~\ref{tab:2} and combining three losses performs the best. We have tested several  mixes too, combining: (i) adversarial \& image space; (ii) adversarial \& ranking; and (iii) image space \& ranking losses. As shown also in~\cite{dosovitskiy2,nguyen}, removing the adversarial objective results to blurry images, and also drops the quantitative and qualitative performances too.

In Table~\ref{tab:3}, we show the ablation study of the impact of different losses on weakly supervised object detection performance. We can clearly observe that, Table~\ref{tab:2} relates very well to that of its individual performance of image reconstruction quantitatively shown in Table~\ref{tab:3}. We observe that the best case in the image reconstruction performs the best in the detection too.

\begin{table}[t]
{\small
\tabcolsep=0.3cm
\caption{Evaluation for the ablation study on losses by quantitative results.}
 \label{tab:2}
\begin{center}
\begin{tabular}{lcc}
\toprule
Method & SSIM & RMSE \\
\midrule
Ours (adv \& img \& ranking loss) & \textbf{0.25} & \textbf{0.30}   \\
Ours (adv \& img loss) & 0.21 & 0.34   \\
Ours (adv \& ranking loss) & 0.23 & 0.32   \\
Ours (img \& ranking loss) & 0.19 & 0.39   \\
\bottomrule
\end{tabular}
\end{center}}

\end{table}

\begin{table}[t]
{\small
\tabcolsep=0.3cm
\caption{The ablation study on losses for object detection performance in mean average precision~(\%).}
\label{tab:3}
\begin{center}
\begin{tabular}{lc}
\toprule
Method & VOC 2007\\
\midrule
Ours (adv \& feature \& ranking loss) & \textbf{45.3}    \\
Ours (adv \& feature loss) & 40.1    \\
Ours (adv \& ranking loss) &  43.7   \\
Ours (feature \& ranking loss) & 34.2   \\
\bottomrule
\end{tabular}
\end{center}}
\vspace{-0.8cm}
\end{table}

\subsection{Weakly supervised detection and classification}
\paragraph*{Comparison with the state-of-the-art:} In this part the weakly supervised detection performance of our generative model is evaluated. Different methods with deep learning~\cite{bilen16,li16}, clustering~\cite{bilen15} and multiple instance learning~\cite{cinbis} approaches are compared with our work. Once our model is trained, it produces the templates for each object instance and the corresponding object heat-maps are extracted via template matching. Using these maps, each object bounding box can be retrieved by looking for the maximum score based on a connected-component method like that of~\cite{gap}. After finding the boxes, Fast-RCNN or SSD object detectors is trained based on our pseudo ground-truth boxes. We report the results for both.  

Table~\ref{tab:4} and Table~\ref{tab:5} present average precision results on PASCAL VOC 2007, 2010 and MS-COCO for object detection. As can be seen, our method outperforms others for this task that use different methods. The {CorLoc} localization performance is also shown in Table~\ref{tab:6}, for PASCAL VOC 2007. Our best performances are 45.3\% and 46.4 (with extra synthesized objects) which is achieved with VGG-16 as the encoder. When using AlexNet, our approach works better than other methods using this network. As to the object proposals EdgeBox~\cite{edgebox} and SelectiveSearch~\cite{SS} are compared based on the Fast-RCNN detector trained by our boxes, we found that Edgebox performs better.

\begin{table}[t]
{\small
\tabcolsep=0.3cm
\caption{Weakly supervised object detection performance in average precision~(\%) comparison on the VOC 2007, 2010, and COCO test set.}
 \label{tab:4}
\begin{center}
\begin{tabular}{lccc}
\toprule
Method &  VOC2007 & VOC2010 & MS-COCO \\
\midrule
Cinbis et al.~\cite{cinbis} &30.2& 27.4 & $-$ \\
Wang et al.~\cite{wang14}& 30.9& $-$& $-$   \\
Li et al., AlexNet~\cite{li16}& 31& 21.4 & $-$  \\
Li et al., VGG16~\cite{li16} & 39.5& 30.7 & $-$  \\
WSDDN~\cite{bilen16} & 39.3& 36.2 & 11.5   \\
WCCN\_AlexNet~\cite{diba} & 37.3 & $-$ & 10.1 \\
Jie et al~\cite{jieSelf} & {41.7} & {38.3} & $-$\\
WCCN\_VGG16~\cite{diba} & 42.8 & 39.5 & 12.3  \\
\midrule
Ours\_AlexNet (FRCNN)  & 39.3 & 38.1 & 10.9 \\
Ours\_VGG (FRCNN) & 44.3 & {41.5} &  12.8 \\
\textbf{Ours\_VGG (SSD)} & \textbf{45.4}& \textbf{43.2}  &  \textbf{13.6} \\
\bottomrule
\end{tabular}
\end{center}}
\vspace{-0.5cm}
\end{table}

Table~\ref{tab:5} shows the effect of adding synthesized samples to a weakly supervised object detection. For each object sample found, we add one synthesized instance to increase the data size for training. With this extra augmentation method the results are improved by 1.1\% and achieved 46.4\%.

\textbf{Object classification:} The proposed method is also tested for classification. The classification results for PASCAL VOC'07 are given in Table~\ref{tab:6}. Once the training for the generative model is completed, the encoder network is fine-tuned for classification with pseudo ground-truth boxes. The main competitor to our work is~\cite{diba}, which used cascaded CNNs to train an object detector and classifier. We can clearly see that our proposed method shows improvement over~\cite{diba}.


\begin{table*}[htb]
{\small
\tabcolsep=0.1cm
\caption{Weakly supervised object detection average precision (\%) on the \textbf{PASCAL VOC 2007} dataset test set.}
 \label{tab:5}
\begin{center}
  \resizebox{\textwidth}{!}{
  \begin{tabular}{  l  c c c c c c c c c c c c c c c c c c c c | c } 
\toprule 
 Method & aero & bike & bird & boat & bottle & bus & car & cat & chair & cow & table & dog & horse & mbike & person & plant & sheep & sofa & train & tv & mAP\\ [0.5ex]
\midrule

Bilen et al. \cite{bilen15} & 46.2 &46.9 &24.1 &16.4 &12.2 &42.2 &47.1 &35.2 &7.8 &28.3 &12.7 &21.5 &30.1 &42.4 &7.8 &20.0 &26.8 &20.8 &35.8 &29.6 &27.7\\ [0.5ex]

Cinbis et al. \cite{cinbis} & 39.3 &43.0 &28.8 &20.4 &8.0 &45.5 &47.9 &22.1 &8.4 &33.5 &23.6 &29.2 &38.5 &47.9 &20.3 &20.0 &35.8 &30.8 &41.0 &20.1 &30.2\\[0.5ex]

Wang et al. \cite{wang14} &48.8 &41.0 &23.6 &12.1 &11.1 &42.7 &40.9 &35.5 &11.1 &36.6 &18.4 &35.3 &34.8 &51.3 &17.2 &17.4 &26.8 &32.8 &35.1 &45.6 &30.9\\[0.5ex]

Li et al., VGG16 \cite{li16} & 54.5 &47.4 &41.3 &20.8 &17.7 &51.9 &63.5 &46.1 &21.8 &57.1 &22.1 &34.4 &50.5 &61.8 &16.2 &29.9 &40.7 &15.9 &55.3 &40.2 &39.5\\[0.5ex]

WCCN\_AlexNet \cite{diba} & 43.9  &57.6 &34.9&  21.3  &14.7&  64.7  &52.8 &34.2 &6.5& 41.2  &20.5 &33.8&  47.6  &56.8 &12.7 &18.8 &39.6 &46.9&  52.9  &45.1   &37.3\\[0.5ex]

{HCP}~\cite{jieSelf} & 52.2 &47.1 &35.0 &26.7 &15.4 &61.3 &66.0 &54.3 &3.0 &53.6 &24.7 &43.6 &48.4 &65.8 &6.6 &18.8 &51.9 &43.6 &53.6 &62.4&    {41.7}\\[1ex]

{WCCN\_VGG16} \cite{diba} & 49.5  &60.6 &38.6 &29.2&  16.2  &70.8 &56.9 &42.5 &10.9 &44.1 &29.9 &42.2 &47.9 &64.1 &13.8 &23.5 &45.9 &54.1 &60.8 &54.5 &{42.8}\\[1ex]

{SGWSOD}~\cite{SGWSOD} & 48.5 &63.2& 33.2& 31.0& 14.5& 69.4& 61.7& 56.6& 8.5& 41.3& 37.6& 50.0& 54.1& 62.7& 22.9& 20.6& 42.1& 50.7& 54.3& 55.2&   {43.9}\\[1ex]

\midrule
Ours\_AlexNet & 45.7  &58.1 &37.2 &24.8 &19 &64.8 &53.7 &35.2 &9.7  &44.8 &22.6 &33.7 &50.4&  57.8  &15.9 &21.7 &40.8 &48.2 &55.4 &45.8   &39.3\\[0.5ex]

Ours\_VGG & 50.9  & 61.2  & 40.5&   31.4&   21.1&   71.6&   58.1&   42.9&   11.7&   46.4&   30.7&   44.5&   48.3&   64.9&   16.8&   24.8&   47.1&   55.7  & 61.7&   55.8&     44.3\\[0.5ex]

\textbf{Ours\_VGG (SSD)} & 51.7&  62.7& 40.6& 33.8& 22.3& 71.4  &59.8&  43.3& 12.5& 48.1& 32.5& 44.8& 49.1& 64.7& 17.4& 25.8& 48.9& 56.7& 63.5& 57.1&   \textbf{45.3}\\[0.5ex]

\textbf{Ours(+Synthesized data)} & 52.4&  63.8& 41.8& 35.1& 22.9& 72.3& 61.1& 44.7& 13.9& 48.6& 32.9& 46.1& 50.7& 66.3& 18.5& 27  &49.7 &56.9&  64.8& 58.6&   \textbf{46.4}\\[0.5ex]

\bottomrule
\end{tabular}}
\end{center}}
\end{table*}

\begin{table*}[htb]
{\small
\tabcolsep=0.1cm
\caption{Classification average precision (\%) on the \textbf{PASCAL VOC 2007} test set.}
\label{tab:6}
\begin{center}
  \resizebox{\textwidth}{!}{
  \begin{tabular}{  l  c c c c c c c c c c c c c c c c c c c c | c } 
\toprule 
 Method & aero & bike & bird & boat & bottle & bus & car & cat & chair & cow & table & dog & horse & mbike & person & plant & sheep & sofa & train & tv & mAP\\ [0.5ex]
\midrule

WSDDN~\cite{bilen16} & 95.0 & 92.6 & 91.2 & 90.4 & 79.0 & 89.2 & 92.8 & 92.4 &78.5 &90.5 &80.4 &95.1 &91.6 &92.5 &94.7 &82.2 &89.9 &80.3 &93.1 &89.1 &89.0\\[0.5ex]

Oquab et al.~\cite{Oquab14} & 88.5 &81.5 &87.9 &82.0 &47.5& 75.5& 90.1 &87.2& 61.6& 75.7& 67.3& 85.5& 83.5 &80.0 &95.6 &60.8& 76.8 &58.0 &90.4 &77.9 &77.7\\[0.5ex]

SPPnet~\cite{SPPNET} & $-$ &$-$ & $-$& $-$& $-$ &$-$ &$-$ &$-$& $-$ &$-$ &$-$& $-$ &$-$& $-$ &$-$& $-$& $-$& $-$& $-$& $-$ &82.4\\[0.5ex]

AlexNet~\cite{bilen16} & 95.3 &90.4 &92.5 &89.6 &54.4 &81.9 &91.5 &91.9 &64.1 &76.3 &74.9 &89.7 &92.2 &86.9 &95.2 &60.7 &82.9 &68.0 &95.5 &74.4 &82.4\\[0.5ex]

VGG16-net \cite{vgg} & $-$ &$-$ & $-$& $-$& $-$ &$-$ &$-$ &$-$& $-$ &$-$ &$-$& $-$ &$-$& $-$ &$-$& $-$& $-$& $-$& $-$& $-$ &89.3\\[0.5ex]

WCCN\_AlexNet~\cite{diba} & 93.1  &91.1 &89.6 &88.9 &81 &89.6 &90.7 &91.2&  76.4  &89.2 &80.8 &92.2 &90.1&  89  &92.7 &82 &89.3&  78.1  &92.8&89.1  & 87.8\\[0.5ex]

{WCCN\_VGG16}~\cite{diba} & 94.2  &94.8&  92.8& 91.7& 84.1& 93  &93.5&  93.9& 80.7& 91.9& 85.3& 97.5& 93.4  &92.6&  96.1& 84.2& 91.1& 83.3& 95.5& 89.6  & {90.9}\\[1ex]

{SGWSOD}~\cite{SGWSOD} & 98.3 & 97.4 & 96.5&  95.7&  79.6&  93.9&  97.5&  96.9&  79.7&  92.3&  82.7&  97.6&  97.2&  95.9&  99.1&  84.2&  92.5&  83.7&  97.3&  92.7&   {92.5}\\[1ex]

\midrule
Ours\_AlexNet & 93.9& 91.8  &90.1&  89.2& 82.7& 89.9  &91.4&  91.6& 77.8& 90.5& 81.7& 92.6& 91  &90.5&  92.8& 83.4& 89.9& 79.6& 93  &90.3&    {88.7}\\[0.5ex]

\textbf{Ours\_VGG16} & 94.8&  95  & 93.2& 91.9& 85  &93.4 &94.1&  94.8& 83  &92.6&  86.7& 97.6& 94  &93.5&  96.4& 85.9& 92.7& 85  &95.7&  90.7&   \textbf{91.8}\\[1ex]

\textbf{Ours(+Synthesized data)} & 94.9 &   95.2&   93.7&   92.7&   85.7&   93.8  & 94.7&   94.7&   84.3&   92.8&   87.5&   98  & 94.5&   93.6&   96.7&   86.4&   93  & 86.8&   96  & 91.1&     \textbf{92.3}\\[1ex]
\bottomrule
\end{tabular}}
\end{center}}
\vspace{-0.1cm}
\end{table*}

\begin{table*}[htb]
{\small
\tabcolsep=0.1cm
\caption{Weakly supervised correct localization (\%) on \textbf{PASCAL VOC 2007} on positive (CorLoc) trainval set.}
 \label{tab:7}
\begin{center}
  \resizebox{\textwidth}{!}{
  \begin{tabular}{  l  c c c c c c c c c c c c c c c c c c c c | c } 
\toprule 
 Method & aero & bike & bird & boat & bottle & bus & car & cat & chair & cow & table & dog & horse & mbike & person & plant & sheep & sofa & train & tv & mAP\\ 
\midrule

Bilen et al.~\cite{bilen15} & 66.4 &59.3 &42.7 &20.4 &21.3& 63.4& 74.3& 59.6& 21.1 &58.2 &14.0 &38.5 &49.5 &60.0 &19.8& 39.2 &41.7 &30.1 &50.2 &44.1 &43.7\\ [0.5ex]

Cinbis et al.~\cite{cinbis} & 65.3 &55.0& 52.4& 48.3 &18.2 &66.4 &77.8 &35.6& 26.5 &67.0 &46.9 &48.4 &70.5 &69.1& 35.2 &35.2 &69.6 &43.4 &64.6 &43.7 &52.0\\[0.5ex]

Wang et al.~\cite{wang14} &80.1& 63.9& 51.5& 14.9& 21.0& 55.7& 74.2 &43.5 &26.2 &53.4 &16.3 &56.7 &58.3& 69.5 &14.1 &38.3 &58.8 &47.2 &49.1 &60.9 &48.5\\[0.5ex]

Li et al., AlexNet~\cite{li16} & 77.3 &62.6& 53.3& 41.4 &28.7 &58.6& 76.2 &61.1 &24.5 &59.6 &18.0 &49.9 &56.8& 71.4 &20.9 &44.5& 59.4& 22.3& 60.9& 48.8 &49.8\\[0.5ex]

Li et al., VGG16~\cite{li16} & 78.2 &67.1& 61.8 &38.1& 36.1& 61.8 &78.8 &55.2& 28.5 &68.8 &18.5& 49.2 &64.1 &73.5& 21.4 &47.4& 64.6 &22.3 &60.9 &52.3 &52.4\\[0.5ex]

WSDDN~\cite{bilen16} & 65.1 &63.4& 59.7 &45.9 &38.5 &69.4 &77.0& 50.7 &30.1 &68.8 &34.0 &37.3& 61.0 &82.9& 25.1 &42.9& 79.2 &59.4& 68.2 &64.1 &56.1\\[0.5ex]

WCCN\_AlexNet~\cite{diba} & 79.7  &68.1&  60.4& 38.9& 36.8& 61.1& 78.6& 56.7& 27.8& 67.7  &20.3 &48.1&  63.9  &75.1 &21.5 &46.9 &64.8 &23.4 &60.2&  52.4  & 52.6\\[0.5ex]

{WCCN\_VGG16}~\cite{diba} & 83.9& 72.8  &64.5&  44.1& 40.1  &65.7 &82.5 &58.9 &33.7 &72.5&  25.6  &53.7&  67.4  &77.4 &26.8 &49.1 &68.1 &27.9 &64.5 &55.7&    {56.7}\\[1ex]

\midrule

Ours\_AlexNet & 83.5  &70.9&  65.4& 42.4& 39  &63.9&  80.8& 58.6& 30.2& 69.5& 24.8& 51  &66.2&  78.4& 25.2& 48.7& 66.6& 26.7& 63.3& 55.9&   55.6\\[0.5ex]

\textbf{Ours\_VGG16} & 85.5 & 75   &66.9 &  47.5 &  43.6 &  67.4 &  83.6 &  61.7 &  36.8 &  75.1 &  29.8 &  55.9 &  70.4 &  80.6 &  29   &52.9 &  71   &31.2 &  66.9 &  58.1 &    \textbf{59.4}\\[1ex]

\bottomrule
\end{tabular}}
\end{center}}
\vspace{-0.1cm}
\end{table*}

\section{Conclusions}\label{sec:conc}
Our novel encoder-generator network contributes to solving important problems in visual recognition, like object discovery and weakly supervised detection. Synthesizing a particular object among the other objects present in an image has been rarely touched upon before. We propose to use different configurations of objective functions to train the visual encoder and generative network, and to utilize the resulting pipeline to separate the objects in an image from each other and the rest of the image. Our approach is a new method to synthesize object categories and to exploit that capability for object detection. 

Combining supervision at the level of object locations and category labels with a ranking hypothesis was shown to be beneficial for training. We have demonstrated the power of our method through different experiments on the PASCAL VOC and MS-COCO datasets. Another important opportunity offered by our method is to train weakly supervised object detectors with the help of object discovery. In future work, we plan to use our proposed pipeline to annotate object bounding boxes for large-scale datasets, such as ImageNet.

\bibliographystyle{splncs}
\bibliography{egbib}
\end{document}